\documentclass{article}

\PassOptionsToPackage{numbers, compress}{natbib}
\usepackage[nonatbib,preprint]{neurips_2023}

\usepackage[utf8]{inputenc} % allow utf-8 input
\usepackage[T1]{fontenc}    % use 8-bit T1 fonts
\usepackage{hyperref}       % hyperlinks
\usepackage{url}            % simple URL typesetting
\usepackage{booktabs}       % professional-quality tables
\usepackage{amsfonts}       % blackboard math symbols
\usepackage{nicefrac}       % compact symbols for 1/2, etc.
\usepackage{microtype}      % microtypography
\usepackage{graphicx}
\usepackage[table,xcdraw, dvipsnames]{xcolor}
\usepackage{multirow}
\usepackage{amsmath}

\usepackage{wrapfig}
\usepackage{subfigure}
\usepackage{caption}

\usepackage[american]{babel}
\usepackage{microtype}

\usepackage{enumitem}

\title{UniParser: Multi-Human Parsing with \\ Unified Correlation Representation Learning}
\author{
  Jiaming Chu \\
  School of Electronic Engineering\\
  Beijing University of Posts and \\
  Telecommunications \\
  Haidian, Beijing \\
  \texttt{chujiaming886@bupt.edu.cn} \\
  \And
  Lei Jin \\
  School of Electronic Engineering\\
  Beijing University of Posts and \\
  Telecommunications \\
  Haidian, Beijing \\
  \texttt{jinlei@bupt.edu.cn} \\
  \And
  Junliang Xing \\
  Tsinghua University \\
  Haidian, Beijing \\
  \texttt{jlxing@tsinghua.edu.cn} \\
   \And
  Jian Zhao \\
  EVOL Lab, Institute of AI (TeleAI), China Telecom\\
  School of Artificial Intelligence, Optics and Electronics\\ (iOPEN), NWPU, Xi'an China \\
  \texttt{zhaoj90@chinatelecom.cn} \\
}

\begin{document}

\maketitle

\begin{abstract}
Multi-human parsing is an image segmentation task necessitating both instance-level and fine-grained category-level information. However, prior research has typically processed these two types of information through separate branches and distinct output formats, leading to inefficient and redundant frameworks. This paper introduces UniParser, which integrates instance-level and category-level representations in three key aspects: 1) we propose a unified correlation representation learning approach, allowing our network to learn instance and category features within the cosine space; 2) we unify the form of outputs of each modules as pixel-level segmentation results while supervising instance and category features using a homogeneous label accompanied by an auxiliary loss; and 3) we design a joint optimization procedure to fuse instance and category representations. By virtual of unifying instance-level and category-level output, UniParser circumvents manually designed post-processing techniques and surpasses state-of-the-art methods, achieving 49.3\% AP on MHPv2.0 and 60.4\% AP on CIHP. We will release our source code, pretrained models, and online demos to facilitate future studies.

\end{abstract}

\section{Introduction}
% 点乘方法改fusion module， 后续补充相关消融实验，以突出task—specific。加上loss function联合优化相关信息，添加是否联合优化消融实验。
Multi-Human Parsing (MHP)~\cite{TianfeiZhou2021DifferentiableMH, LuYang2018ParsingRF, TaoRuan2019DevilIT, JianshuLi2017TowardsRW, FangtingXia2017JointMP} aims to segment fine-grained human body parts for each person in an image, which requires distinguishing both instance-level persons~\cite{DanielBolya2019YOLACTRI, EnzeXie2020PolarMaskSS} and category-level body parts~\cite{AlirezaFathi2017SemanticIS, WenbinHe2022SelfsupervisedSS}. With the widespread adoption of deep learning techniques, MHP has made remarkable progress and played a crucial role in various applications, including virtual try-on~\cite{2017Body, 2018Virtual}, human-computer interaction~\cite{2013Human, 2021Research}, and scenario understanding~\cite{Yongkang2015Deep, 2014User}, \emph{etc}.

\begin{figure}[t]
    \vspace{-5mm}
    \centering    \includegraphics[width=1\columnwidth]{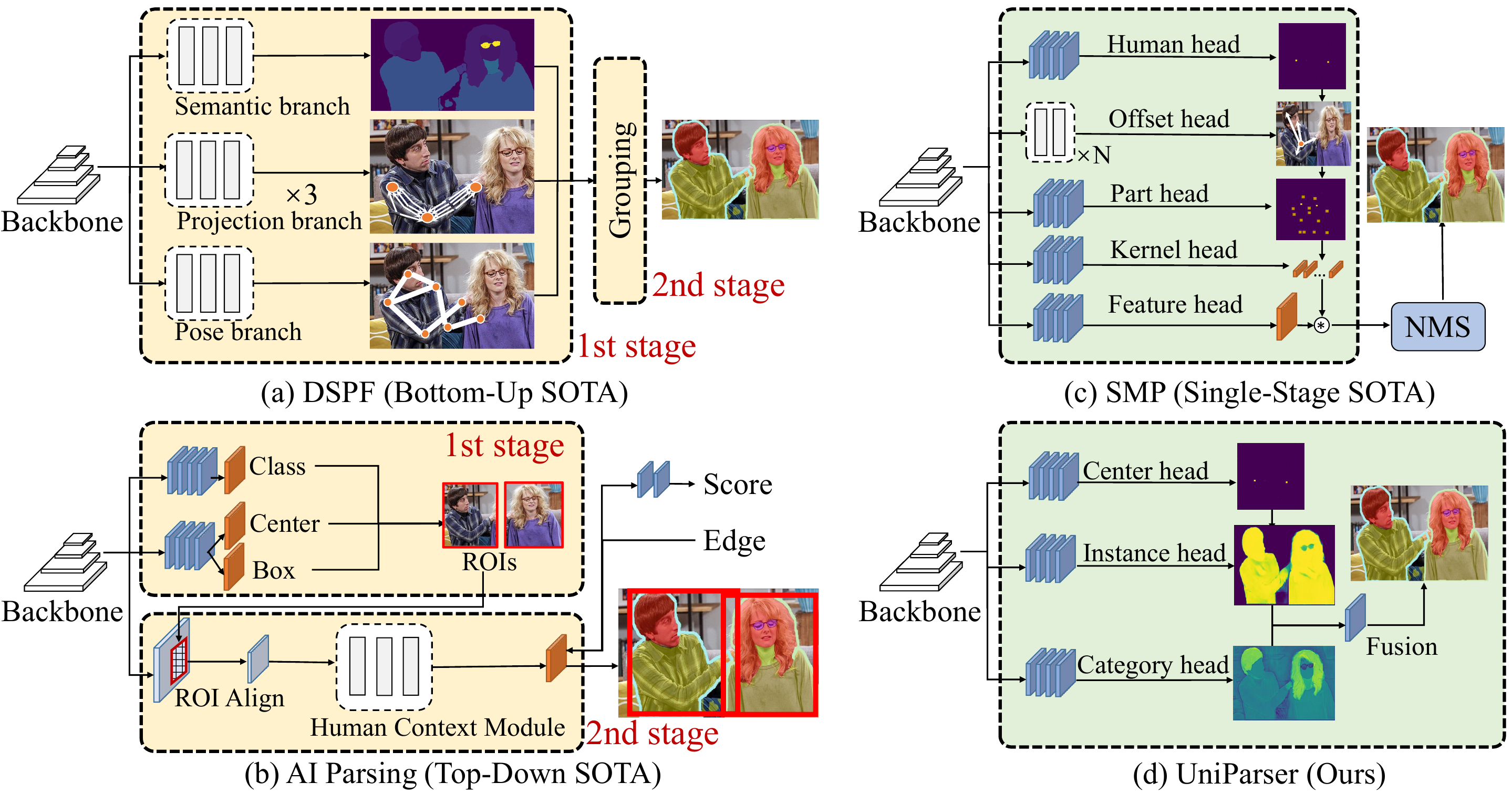}
    \vspace{-1mm}
    \caption{Illustration of state-of-the-art network architectures for multi-human parsing. (a) DSPF~\cite{TianfeiZhou2021DifferentiableMH}; (b) AIParsing~\cite{SanyiZhang2022AIParsingAI}; (c) SMP~\cite{2023SMP}; (d) UniParser (Ours). \textcolor{Dandelion}{\textbf{Yellow}}: two-stage; \textcolor{LimeGreen}{\textbf{Green}}: single-stage.}%标题中使用一两句话点出UniParser的显著特点
   
    \label{fig.Difference}
    \vspace{-5mm}
\end{figure}

The current MHP methodologies can be broadly classified into three categories: bottom-up, top-down, and single-stage methods. As shown in Fig.~\ref{fig.Difference} (a), the bottom-up methods~\cite{JianZhao2020FineGrainedMP, JianshuLi2021MultihumanPW, TianfeiZhou2021DifferentiableMH}, first parse all human parts in the image, and then group them into corresponding instances based on grouping clues, \emph{e.g.}, human edge, pose, \emph{etc}. The top-down methods~\cite{KaimingHe2017MaskR, LuYang2018ParsingRF, SanyiZhang2022AIParsingAI}, as shown in Fig.~\ref{fig.Difference} (b), initially employs a detector to identify human instances and subsequently perform individual parsing of human parts for each instance. The single-stage methods~\cite{2023SMP, 2022RepParser}, as shown in Fig.~\ref{fig.Difference} (c), predict human instances and part masks in parallel and usually need an NMS process~\cite{XinlongWang2020SOLOv2DA} to obtain final results. Overall, the bounding box in the top-down method, grouping clue in bottom-up methods, and center-based offsets in single-stage methods are instance-level information. The fine-grained semantic segmentation for human parts is category-level information. Despite the superior performance of previous research, they separately  process instance and category information in different forms. Furthermore, due to the presence of detectors, grouping clues, and NMS processes, their entire pipeline is not jointly optimized nor truly end-to-end. As a result, previous research pipelines are inefficient and redundant without direct supervision of final results and a purely end-to-end process. 

To address the issues of separating instance and category, we propose UniParser as an end-to-end framework that unifies instance-level and category-level representations without post-processing. Firstly, we put forward a unified correlation representation learning in the cosine space to extract instance and category features. Specifically, we model the instance feature by correlations between different instances and the category feature by correlations between predefined category parameters and images across the entire dataset. Then we unify the network output format as pixel-level segmentation results, and instance and category features are intermediately supervised by corresponding ground-truth masks with an auxiliary loss. Finally, we integrate the instance and category features through a fusion module, in which the instance-aware human part parsing can jointly optimize the two features. As shown in Fig.~\ref{fig.Difference} (d), our pipeline keeps unified format and is optimized jointly, resulting in higher efficiency and a more compact design than other representative methods (see Sec.~\ref{sec:4.2}). %Our source code, pretrained models, and online demos will be released upon acceptance.

The main contributions can be summarized into three aspects as follows:
\vspace{-1mm}
\begin{itemize}[leftmargin=0cm, itemindent=0.3cm, itemsep=1.2mm]
\item To the best of our knowledge, we are the first to propose to unify instance and category representations, and optimize them jointly to solve fine-grained parsing tasks. It enables mutual benefits between the two representations and realizes the simplest solution.
\item We further introduce correlation representation learning to facilitate our network acquiring instance and category features. Initially, we map the feature vectors by normalization to cosine space. Subsequently, we utilize correlations between instances and correlations within and between categories to represent instance and category features, respectively. As a result, both types of features are trained uniformly in a more distinguishable manner within their respective categories or instances.

\item With the above two proposed guidelines, we construct a purely end-to-end MHP framework, termed as UniParser, which eliminates the need for complex post-processing and directly produces human parsing results. Without bells and whistles, UniParser outperforms all other state-of-the-art methods on MHPv2.0 and CIHP with the fastest inference speed.
\end{itemize}

\section{Related Work}
This section provides a comprehensive review of relevant research on our methods, including top-down, bottom-up, and single-stage multi-human parsing methods. 

\textbf{Top-down Methods.}
The top-down methods typically rely on detectors to locate bounding boxes of instances, followed by single-person parsing within the detected regions. Parsing RCNN~\cite{LuYang2018ParsingRF} and RP-RCNN~\cite{LuYang2020RenovatingPR} are classical researches, generating a bounding box first and utilizing an additional context encoding module to get finer semantic pixel features. There are also some methods~\cite{TaoRuan2019DevilIT, RuyiJi2019LearningSN, SanyiZhang2022AIParsingAI} focusing on mining new constraints to enhance the ability of the model to extract instance context features. %SNT~\cite{RuyiJi2019LearningSN} relies on Mask RCNN to distinguish human instances, predicts human parts by attributes in stages, and can reduce prediction difficulty. 
However, top-down methods consist of an independent detector and a single-person parser, leading to non-end-to-end optimization. 

\textbf{Bottom-up Methods.}
The bottom-up methods typically involve initially predicting the semantic label of pixels~\cite{HexinDong2022RegionAwareML,WenbinHe2022SelfsupervisedSS,JieLi2021IMENetJ3}, followed by grouping them based on instance information. Some works~\cite{JianZhao2020FineGrainedMP,JianshuLi2017TowardsRW} apply the GAN-based network to improve the learning ability of instances and semantic features of the model. Other works utilize additional ground-truth information such as pose~\cite{TianfeiZhou2021DifferentiableMH} and edge~\cite{KeGong2018InstancelevelHP} to help the model learn instance characteristics. However, in bottom-up methods, instance and semantic information are processed independently, necessitating an additional grouping step.

\textbf{Single-stage Methods.}
The single-stage methods~\cite{XuechengNie2019SingleStageMP,ZigangGeng2021BottomUpHP,JinLei2022SingleStageIE} aim to accomplish the task in a single pass without involving ROI detection and grouping process. Currently, only a few single-stage methods are available, such as SMP~\cite{2023SMP} and RepParser~\cite{2022RepParser}. SMP refers to the single-stage method in pose estimation and models the multi-human parsing task by point sets and offsets, generating human parsing by conditional kernels. While RepParser decouples the parsing pipeline into instance-aware kernel generation and part-aware human parsing. Although both methods have advantages in speed, they still process instance and category information in different branches and need non-differentiable post-processes during inference. In our framework, we propose integrating instance and category representations and eliminating post-processing through correlation representation learning, achieving a purely end-to-end optimization. %We rethink the representation of the task, and attempt to establish a simple but effective single-stage method.

\section{UniParser}
\subsection{Overview}
\textbf{Task Description.}
\label{task description}
Given an image $I$, multi-human parsing task aims to parse the human instances in $I$ and obtain their masks of body parts $\small \mathcal{M} = {\{M^{human}_{m}\}}^{N_{ins}}_{m=1}$, where $N_{ins}$ denotes the number of human instances in $I$. For each human instance, it is composed of part instance segmentation  $\small M^{human}_{m} = {\textstyle \sum_{i=1}^{N_{cate}} M^{part}_i }$, where $N_{cate}$ is the number of part categories, $M^{part}_i$ is the mask of $i$-th class part. From the above description, we can simplify the multi-human parsing task by assigning a human instance label $m \in N_{ins}$ and parsing label $c \in N_{cate}$ to each pixel on the image.

\begin{figure}[t]
    \vspace{-4mm}
    \centering
    \includegraphics[width=1\columnwidth]{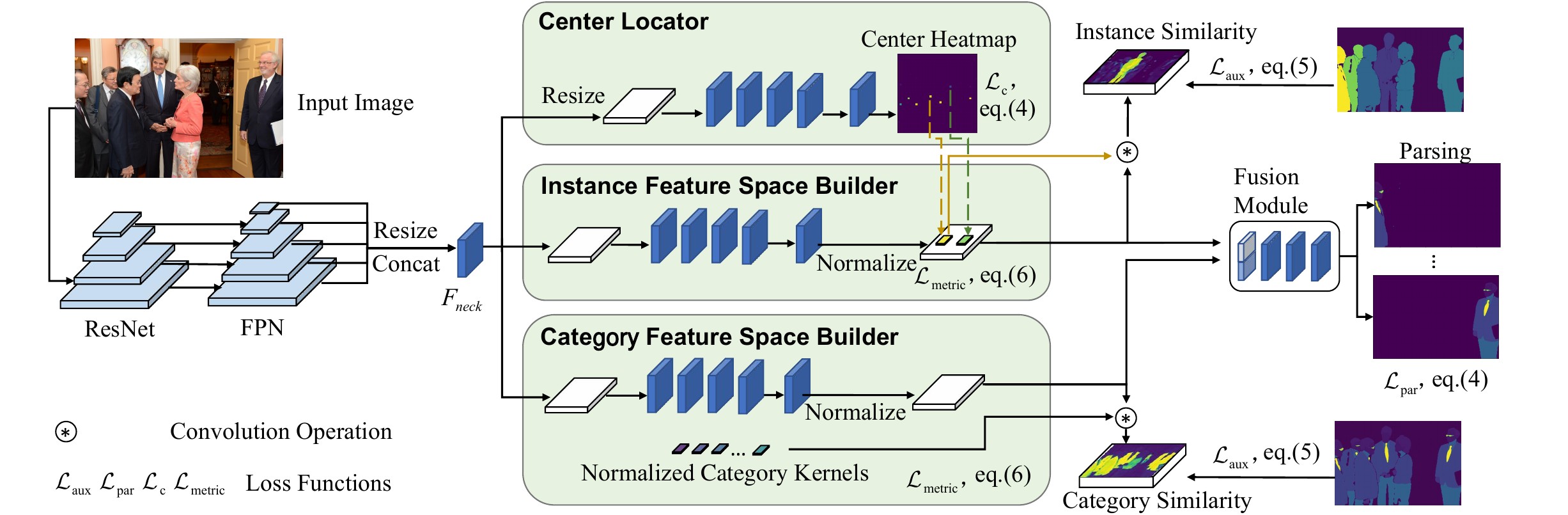}
    \caption{\textbf{Overview of UniParser.} ResNet and FPN are utilized to obtain multi-scale features sent to the following modules. The Center Locator is responsible for localizing human barycenters. Instance Feature Space Builder and Category Feature Space Builder learn instance and category correlation representations in cosine space, respectively. The fusion module combines instance and category features to predict instance-aware human parsing.
    }
    %An input image is first sent into backbone with FPN to obtain multi-scale features, which serve as input for Center Locator (CL), Instance Feature Space Builder (IFSB), Category Feature Space Builder (CFSB).
    \vspace{-4mm}
    \label{fig.Framework}
\end{figure}

\textbf{Overall Architecture.}
%Our framework models MHP tasks in a compact way. As shown in Fig.~\ref{fig.Framework}, UniParser consists of four main components, \emph{i.e.}, backbone, Center Locator~(CL), Instance Feature Space Builder~(IFSB), and Category Feature Space Builder~(CFSB).  CL and IFSB aim to learn instance-level representation in cosine space with spatial feature mining. CFSB generates category-level representation in cosine space with a set of category parameters trained on the entire dataset. Finally, we fuse outputs from IFSB and CFSB with a fusion module to obtain fine-grained human parsing. In this way, IFSB and CFSB can be optimized jointly. 
Our framework models the MHP task in a compact way. To enhance the extraction of instance features, we propose instance-level correlation representation learning, which includes Center Locator and Instance Feature Space Builder as illustrated in Fig.~\ref{fig.Framework}. Moreover, we propose category-level correlation representation learning to acquire category-specific features with Category Feature Space Builder, as illustrated in Fig.~\ref{fig.Framework}. Finally, we fuse outputs from the above two branches with a fusion module to obtain fine-grained human parsing. %In this way, instance-level and category-level features can be optimized jointly. 

%The instance branch consists of a barycenter localization sub-branch and a instance mask generation sub-branch. The barycenter localization sub-branch is used to locate the center of the gravity position of instance in the image. On the basis of determining the barycenters, we utilize self-convolution and metric loss function in instance mask generation sub-branch to increase the similarity of pixel features belonging to the same instance in the cosine space, and vice versa. We regard cosine similarity as a confidence to obtain the mask of the human instance. The semantic branch utilizes the metric loss function to decrease the cosine similarity between the convolution kernels, and generates semantic masks through convolutional operations. Finally, our method can directly output fine-grained human parsing by multiplying outputs of instance branch and semantic branch. Furthermore, SP is CNN-based and end-to-end, without any post-processes.  

%\noindent\textbf{Backbone \& Neck.}
%\textbf{Backbone \& Neck.}
In UniParser, we select the relevant models of the ResNet~\cite{KaimingHe2015DeepRL} as backbones for feature extraction. To simultaneously learn instance and scale features in different scales, we adopt FPN~\cite{TsungYiLin2016FeaturePN} to fuse features. We concatenate the outputs of each layer of the FPN together and then use a convolution layer of 1 $\times$ 1 to perform feature compression to obtain a map containing multi-scale features $F_{neck}$.

%\subsection{Instance-level Representation Learning in Cosine Space}
\subsection{Instance-level Correlation Representation Learning}
We propose Center Locator (CL) and Instance Feature Space Builder (IFSB) to facilitate instance-level feature extraction, which enables differentiation between individuals in an image. Given that recognizing different instances requires a more robust discriminative ability than semantic segmentation, we opt for the dynamic distribution of instance features based on their positions. Consequently, we design CL and IFSB as follows.%Due to the uncertainty in the number of instances in an image and the uniqueness between instances; their features should be dynamically distributed according to instance positions. In order to comply with the above characteristic, we design CL and IFSB as follows. 

\textbf{Center Locator~(CL)} aims to locate the barycenter of each human instance mask, as the instance information is strongly correlated with the spatial center of each instance. Given the sparsity of the barycenter, large feature maps are not necessary for accurate prediction. Therefore, before entering the branch, the output feature $F_{neck}$ of FPN is resized to a fixed scale size $\small S\times S$ ($S=40$) by bi-linear interpolation, and the relative coordinate is concatenated. After the processed feature passes through 5 convolution layers with $3 \times 3$ kernels, the final output is a heatmap $H_c$ with a shape of $\small S \times S \times 1$ that represents the confidence of the human barycenter.
The process is as follows:
\begin{equation}\label{eq1}
\small H_{c}^{1\times S\times S} = \mathbf{Sigmoid}(\mathbf{g_{CL}}(\mathbf{Resize}(F_{neck}^{C\times H\times W}, (S, S)))), 
\end{equation}
where $\mathbf{g_{CL}(\cdot)}$ represents stacked convolution layers. A sample output of CL is shown in Fig.~\ref{fig.intermediate_results} (a).

\textbf{Instance Feature Space Builder~(IFSB)} aims to cluster pixel features belonging to the same instance in cosine space while simultaneously separating them from features of other instances and backgrounds. To generate an elaborate instance feature, IFSB does not resize the feature map $F_{neck}$, but directly sends it into five convolution layers with $3 \times 3$ kernels after concatenating the relative coordinates. Then extracted features are normalized along the channel dimension, with each pixel feature having a length of 1. By utilizing the prediction results from CL, we can obtain the barycenter coordinates and select corresponding pixel features $f_{x,y}$ in the normalized feature map $F_{ins}$. We regard these features as convolution kernels with the normalized feature map to generate cosine similarity maps. The value of each pixel in the cosine similarity map represents the cosine similarity between the convolution kernel $f_{x_i,y_i}$ and other pixel features in the feature map. 
The above process could be summarized as follows:
\begin{equation}
\small
F_{ins}^{C\times H\times W} = \mathbf{Normalize}\left(\mathbf{g_{IFSB}}\left(F_{neck}^{C\times H\times W}\right)\right), \qquad \left\{ \left(x_i,y_i\right) \right\} = \mathbf{WHERE}\left(H_{c}^{1\times S \times S}>\theta_{c}\right), \nonumber
\end{equation}
\begin{equation}
 \small  f_{x_i,y_i} = F_{ins}\left[x_i,y_i\right], \quad i \in \{1...,N_c\}, \qquad \quad   Q_{ins}^{N_c \times H \times W} = \mathbf{Concat}\left(f_{x,y}\right)^{N_c \times C \times 1 \times 1} \ast F_{ins}^{C \times H \times W},  
\end{equation}
\vspace{0.3em}
where $\mathbf{g_{IFSB}(\cdot)}$ denotes the function of multiple convolution layers in IFSB, $F_{ins}$ is the mapped vectors in features space trained by IFSB, \textbf{WHERE} function aims to index the coordinates of whose value is over $\theta_c$ (we set $\theta_c=0.1$ in experiments), $N_c$ is the number of above-indexed coordinates. $Q_{ins}$ is the collection of cosine similarity maps of instance features.
During training, guided by a loss function, the cosine similarity between pixel features belonging to the same instance will become larger. As shown in Fig.~\ref{fig.Cosine_space}~(b), $\theta_{inter}$ will be larger and $\theta_{intra}$ will be smaller during training. All pixel features whose $\theta_{intra}$ are smaller than the threshold would be classified into one instance. For details of the loss function for IFSB, please refer to Sec.~\ref{Loss function}. Visualization examples for similarity maps in IFSB are illustrated in Fig.~\ref{fig.intermediate_results} (b)(c).

%\subsection{Category-level Representation Learning in Cosine Space}
\subsection{Category-level Correlation Representation Learning}
We propose the Category Feature Space Builder~(CFSB) to learn category-level features in cosine space. Unlike instance features, which may vary across instances, category features remain consistent as they are based on general category-level information independent of specific instances. Thus, we do not need to dynamically adjust the position of each category in the cosine space based on different inputs. We just set a fixed position for each category in cosine space.

\begin{figure}
    \vspace{-4mm}
    \centering
    \includegraphics[width=0.95\columnwidth]{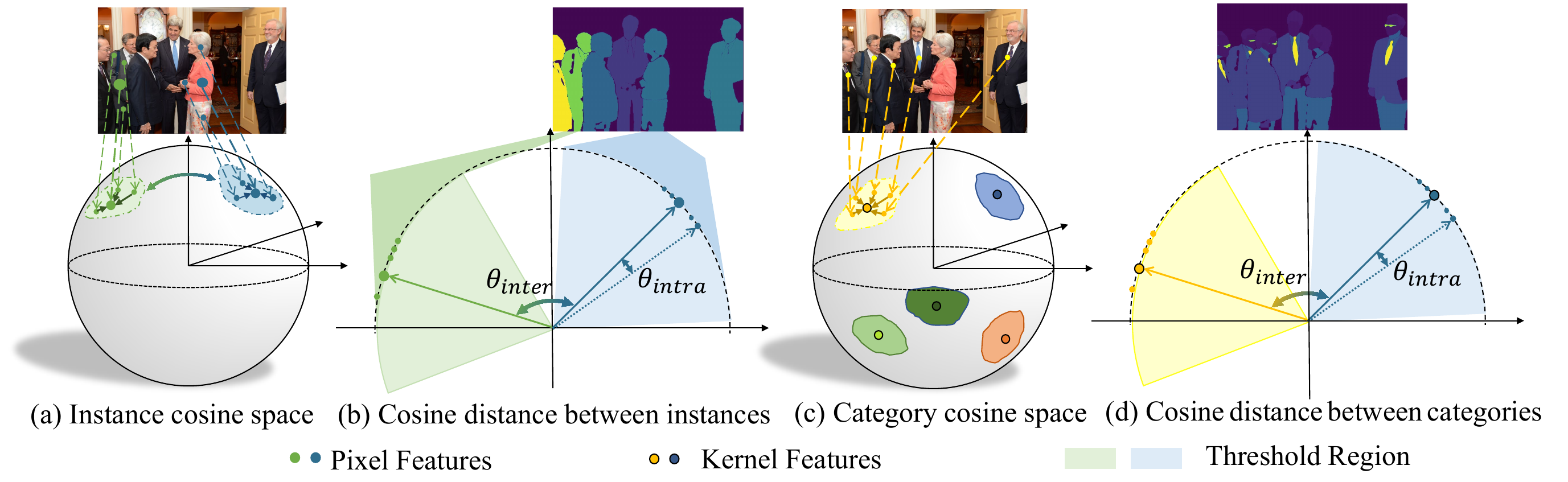}
    \caption{Cosine space trained for instance and category feature. $\theta_{inter}$ denotes the cosine distance between different instances or categories. $\theta_{intra}$ denotes the cosine distance between pixel features in the same instance or category.}
    \vspace{-5mm}
    \label{fig.Cosine_space}
\end{figure}

\textbf{Category Kernels $K_{cate}$} are designed to maintain the consistency of category semantics, serving as anchors of the semantics of each category in the category cosine space. We initialize this group of parameters whose shape is $\small (N_{cate}, C, 1, 1)$, where $N_{cate}$ is the number of classes in the dataset, $C$ is the number of channels of the previous output. Different image pixel features will gradually approach their corresponding category kernel $K_{cate}$ during training, as shown in Fig.~\ref{fig.Cosine_space}~(c). Unlike the instance kernels of IFSB, which the center of human instances dynamically adjusts, category kernels are pre-positioned and remain unchanged across different input images during inference. Additionally, category kernels are also normalized along channels.

\textbf{Category Feature Space Builder~(CFSB)} is specifically designed to pull positive pixel features towards their corresponding category kernel. Firstly, for fine-grained masks, the output of FPN is directly fed into 5 convolution layers without resizing. Additionally, we normalize the features along the channel dimension before convolving them with category kernels to obtain category-level features. This normalization ensures that all pixel category features have a length of 1, and $K_{cate}$ is also normalized. Therefore, their convolution results are their cosine similarity map. The above process could be summarized as follows: 

\begin{equation}
\small
 F_{cate}^{C \times H \times W} =\mathbf{Normalize}\left(\mathbf{g_{CFSB}}\left(F_{neck}\right)\right), \quad  Q_{cate}^{N_{cate} \times H \times W} = K_{cate}^{N_{cate} \times C \times 1 \times 1} \ast F_{cate}^{C \times H \times W},  
\end{equation}

where $\mathbf{g_{CFSB}}$($\cdot$) represents the function of 5 convolution layers, $K_{cate}$ denotes the category kernels, $Q_{cate}$ denotes the cosine similarity maps between the kernels in $K_{cate}$ and the normalized features $F_{cate}$. Similarly, as the instance feature, during training, the category features $F_{cate}$ belonging to the same category will approach the corresponding category kernel in the cosine space, as shown in Fig. ~\ref{fig.Cosine_space} (c)(d). In addition, the category kernels can be trained to explore optimal positions in the cosine space and remain fixed during inference. For details of the loss function for CFSB, please refer to Sec.~\ref{Loss function}. Visualization examples for similarity maps in CFSB are illustrated in Fig.~\ref{fig.intermediate_results} (d)(e).

%We utilize the instance kernels $f_{x,y}$ and category kernels $K_{cate}$ to calculate the cosine similarity maps $Q_{ins}$ and $Q_{cate}$. 
%As shown in Fig.~\ref{fig.Cosine_space}~(b) (d), we generate instance masks $M_{ins}^{N_c \times H \times W}$ and category masks $M_{cate}^{N_{cate} \times H \times W}$ by segmentation with threshold, and the pixels whose similarity  with an instance or a category is smaller than threshold would be classified into a corresponding instance or category.

% \begin{equation}
%  \small  Q_{parsing} = \mathbf{F_{fusion}}(Q_{ins}, Q_{cate}),
% \end{equation}

%Consistent with the instance mask generation sub branch, the output of FPN are directly input into five convolution layers without resizing. During model initialization, the model contains semantic convolution kernels $K_s$ with a shape of $(N_{class}, C, 1, 1)$, where $N_{class}$ is the number of semantic classes, $C$ is the number of channels of semantic branch output. After normalization, the semantic features and the semantic convolution kernels are convolved to obtain cosine similarity map of semantics. Similarly, the cosine similarity between the pixel feature and semantic convolution kernel represents the confidence that the pixel belongs to the semantic category.

%\subsection{Representation Learning in Cosine Space}

\subsection{Training and Inference}

\textbf{Fusion Procedure with Joint Optimization.}
After mapping features into the cosine space, we adopt a fusion module to fuse instance features $F_{ins}$ and category features $F_{cate}$ to accomplish joint optimization, which can be denoted as $\small  Q_{parsing} = \mathbf{g_{fusion}}(F_{ins}, F_{cate})$. Furthermore, we have designed several kinds of fusion strategies (Detailed comparison results can be found in Tab.~\ref{tab:ablation} (d)). Finally, we find that the element-wise indexing performs best.

\textbf{Training Details.}
\label{Loss function}
The overall training loss of UniParser is defined as follows:
\begin{equation}
\small \mathcal{L}_{\rm total} = \mathcal{L}_{\rm c} + \lambda_{\rm aux} \mathcal{L}_{\rm aux} + \lambda_{\rm par} \mathcal{L}_{\rm par} + \lambda_{\rm metric} \mathcal{L}_{\rm metric},
\end{equation}
where $\lambda_{\rm aux},\lambda_{\rm par},\lambda_{\rm metric}$ are parameters to balance different loss items.

The \emph{center loss} $\mathbf{\mathcal{L}_{\rm c}}$ is the loss function for Center Locator. For ground-truth heatmap $H_c^*$, we set values in barycenters of each human instance mask as one and add the values around barycenters as positive samples according to human sizes. Then we utilize Focal loss~\cite{TsungYiLin2017FocalLF} to supervise $H_c$. 
%For the heatmap $H_c$, we generate center map label by resizing the ground-truths of instance masks and calculating their barycenters, and to add generalization of CL, we choose 2-6 pixels around barycenter for each instance to be positive labels. And we utilize Focal Loss to supervise it. 

The \emph{auxiliary loss} $\mathcal{L}_{\rm aux}$ in IFSB and CFSB aims to enhance the proximity between pixel features that belong to the same instance or category, which comprises three components for constructing a cosine feature space. Ideally, pixel features belonging to the same instance, or category should exhibit a similarity score of 1 in cosine space. Therefore, masks representing instances or categories can be ground-truths for measuring such similarity. Each pixel within an instance is assigned a value of 1, while others are set at 0. The equation of $\mathcal{L}_{\rm aux}$ is defined as follows:

\begin{equation}
\small
\mathcal{L}_{\rm aux} = \frac{1}{N_{c}} \sum_i{{\rm Dice}\left( q_{ins}^{i},q_{ins}^{i*} \right)} +\frac{1}{N_{cate}}\sum_{j,k}{{\rm Dice}\left( q_{cate}^{j},q_{cate}^{j*} \right) +\ell _1\left( q_{cate}^{k},0 \right)} ,
\end{equation}
% \begin{equation}
% \small \mathcal{L}_{\rm aux} = \textbf{DiceLoss}(Q_{ins}, Q_{ins}^*) +   \textbf{DiceLoss}(Q_{cate_{pos}}, Q^*_{cate_{pos}}) +  {\mathbf{\ell_1}}(Q_{cate_{neg}}, \textbf{0})
% \end{equation}

where $\small q_{ins}^{i}\in Q_{ins},\ q_{cate}^{j}\in Q_{cate}^{pos},\ q_{cate}^{k}\in Q_{cate}^{neg}$. $Q_{cate}^{pos}, Q_{cate}^{neg}$ are the collections of similarity maps for existent and non-existent categories in an image, respectively. $\small q_{ins}^{i*}$ is the ground-truth mask for human instance $i$. $q_{cate}^{j*}$ is the ground-truth mask for part category $j$.
We adopt Dice Loss~\cite{hcr:FaustoMilletari2016VNetFC} to train the two builders. However, Dice Loss just focuses on the areas whose values are larger than 0, so the model will not learn anything from non-existent categories. We further introduce the $\mathbf{\ell_1}$ loss function to learn more information from non-existent categories in CFSB. Similarity maps of non-existent categories are 0. %We reduce the values of prediction in Non-existent similarity maps by $\mathbf{\ell_1}$ loss function.

The \emph{parsing loss} $\mathcal{L}_{\rm par}$ is designed for final outputs. We choose %$\mathbf{Diceloss}(Q_{parsing}, Q^*_{parsing})$ 
Dice loss~\cite{hcr:FaustoMilletari2016VNetFC} to supervise the instance-aware parsing results and it could jointly optimize IFSB and CFSB.

% \subsubsection{Self-Convolution}
% \label{Self-Convolutio}
% Self convolution refers to the conduct convolution on the feature map with the kernels which are the features from the feature map itself. 
% \begin{equation}
% \small
% Inner_{ins}^{N\times H \times W} = F_{ins}^{C\times H \times W} \otimes f^{N \times C\times 1 \times 1}
% \end{equation}
% where $N$ is the number of instances, $F_{ins}$ is the output of instance mask generation sub-branch, $f$ is the pixel feature indexed by barycenter heatmap $H_c$ from $F_{ins}$.
% Convolutional operation can be regarded as calculating the inner product similarity between each pixel feature in the feature map and the kernel. When the feature map $F_{ins}$ and kernel $f$ are normalized, the convolutional result is equivalent to the cosine feature map.
% \begin{equation}
% \small
% Cos_{ins}^{N\times H \times W} = Normalize (~F_{ins}^{C\times H \times W}) \otimes Normalize(~f^{N \times C\times 1 \times 1})
% \end{equation}

% Under the supervision of the loss function, the similarity between the pixel features belonging to the same instance and the cosine of the convolution kernel will increase, and vice versa. Compared to traditional dynamic convolution, self-convolution does not require additional kernel branch to generate kernels, which come from the feature map itself and enable the model to learn discriminative knowledge more directly. The cosine similarity of self convolution output can be directly used as confidence to obtain mask by threshold. 

%\subsubsection{Metrics loss function}

\textbf{Metric Loss.} $\mathcal{L}_{\rm metric}$ focuses more on the overall diversity between instances or categories than on that between pixels. $\mathcal{L}_{aux}$ and $\mathcal{L}_{\rm parsing}$ are both committed to reducing the $\theta_{intra}$ in Fig.~\ref{fig.Cosine_space}, but $\mathcal{L}_{\rm metric}$ is committed to increasing the $\theta_{inter}$. We calculate the cosine similarity inside $f_{x,y}$ or $K_{cate}$, and get the similarity matrices $A_{ins}$ and $A_{cate}$. Ideally, $f_{x_i,y_i}$ and $K_{cate_j}$, representing different instances or categories, should have a cosine similarity of 0 with each other, and their similarity maps should be a diagonal matrix with values of 1. We define $\mathcal{L}_{\rm metric}$ as follows:

\begin{equation}
 \small \! \mathcal{L}_{\rm metric} = \frac{1}{N_{cate}+N_{c}}(\sum(\mathbf{\ell_1}(A_{cate}^{N_{cate} \times N_{cate}},   {\rm Diag}^{N_{cate} \times N_{cate}})) + \sum(\mathbf{\ell_1}(A_{ins}^{N_c \times N_c}, {\rm Diag}^{N_c \times N_c}))),
\end{equation}

where $\rm Diag$ denotes a diagonal matrix whose values of the diagonal are 1. %$\mathbf{Sum}$ is the summation operation.

\textbf{Inference.} During inference, we set a threshold for human part masks. Then we can directly obtain instance-aware human parsing from outputs of the fusion module $Q_{parsing}$. Due to the unified pipeline and corresponding loss functions, UniParser is purely end-to-end and avoids the Non-Maximum Suppression (NMS) process~\cite{XinlongWang2020SOLOv2DA} or other post-processes.

%Moreover, in UniParser, we have developed a novel method that eliminates the need for NMS. %Hungarian assignment strategy often appear in transformer based models, where they specify a query for each target in the image, thus avoiding redundant predictions. 

\section{Experiments and Analysis}

\subsection{Experimental Setup}

\noindent\textbf{Dataset.} MHPv2.0~\cite{JianZhao2018UnderstandingHI} is the most extensive and demanding dataset in the realm of multi-human parsing. It comprises 15,403 training images and 5,000 validation images. Meanwhile, the dataset also provides labels for 16 keypoints of human pose. %Each image contains between 2 to 26 human instances with an average of 2.6 instances per image. 
It includes a total of 58 part categories consisting of 11 labels for human body parts and 47 labels for clothing and accessories. CIHP~\cite{KeGong2018InstancelevelHP} is a large-scale multi-human parsing dataset in the wild. It comprises 38,280 human instances with 19 pixel-wise human semantic parts and corresponding instance IDs. The dataset contains a total of 28,280 images for training, 5,000 images for validation, and 5,000 images for testing. %with an additional 5,000 images each allocated for validation and testing purposes. 

\noindent\textbf{Metrics.}
For the evaluation of instance-aware human parsing performance, we utilize the Average Precision based on part (AP$^{p}$)~\cite{JianZhao2018UnderstandingHI} for multi-human parsing assessment. This metric calculates the mean part instance-level pixel IoU of different semantic part categories within a human instance to determine if the human instance is a true positive. We choose the AP$^{p}_{50}$ and AP$^{p}_{vol}$ as the evaluation metrics. The former defines instances with an IoU greater than 0.5 as positive, while the latter calculates the average AP$^{p}$ across IoU thresholds from 0.1 to 0.9 in increments of 0.1. In addition, we also present the official metric known as the Percentage of Correctly parsed semantic Parts (PCP)~\cite{JianZhao2018UnderstandingHI}.

\textbf{Implementation Details.} We implement our UniParser using mmdetection~\cite{mmdetection} on a server equipped with 2 NVIDIA 3090 GPUs and 24GB memory per card. We utilize ResNet-50, ResNet-101, and FPN as the backbone and neck in all architectures, each of which is trained end-to-end. A mini-batch involves 8 images. We employ scale jittering by randomly sampling the shorter side of the image from 640 to 800 pixels, following SOLO~\cite{XinlongWang2019SOLOSO}. We regard 12 epochs as a 1$\times$ training schedule with an initial learning rate of 0.005, which is then divided by 10 at the 9-th and 11-th epochs. And 1$\times$ training schedule is adopted in the ablation study. We adopt 3$\times$ training schedule with 36 epochs whose learning rate is divided by 10 at 27-th and 33-th epoch for performance comparison.

\subsection{Main Results}\label{sec:4.2}
\textbf{MHPv2.0 val set.} Tab.~\ref{The main results on MHP} presents the results of comparison with the state-of-the-art top-down and bottom-up methods on MHPv2.0 val set. Our approach significantly outperforms both top-down, bottom-up, and single-stage models in all instance-aware metrics, requiring fewer training epochs and without the need for additional ground truth beyond masks. %For other methods, the longer learning schedule is set up to 150 epochs, whereas our method sets 36 epochs as a 3$\times$ learning schedule. 
With ResNet-101~\cite{KaimingHe2015DeepRL} serving as the backbone, UniParser achieves 51.2\%, 49.3\%, and 52.9\% on AP$^{p}_{50}$, AP$^{p}_{vol}$, and PCP$_{50}$, outperforming by +4.1, +1.1, and +1.4 than SMP~\cite{2023SMP}, which is the existing SOTA model in MHP task. As shown in Fig.~\ref{fig.intermediate_results} (h)(i), UniParser can produce more precise parsing results compared with SMP~\cite{2023SMP}, particularly in smaller parts such as watch, glasses, and so on. We design a lighter UniParser to further explore a faster inference model, denoted as UniParser$_{light}$, which can get comparable performance with the original. Specifically, we select the last stage of $F_{neck}$ as input for the subsequent builders, connecting them with a convolution layer with 1 $\times$ 1 kernels.

\begin{table*}[t]
\vspace{-4mm}
\renewcommand\arraystretch{1.0}
\caption{The comparable results on MHPv2.0 dataset. To facilitate better comparison, we denote the state-of-the-art methods under identical settings in gray. $light$ refers to the reduction of convolution channels and kernel sizes in the feature compression layer after FPN.}
\vspace{-1mm}
\setlength{\tabcolsep}{8.4pt}
\small
\centering
\begin{tabular}{llccccccc}
%\hline
\multicolumn{1}{l|}{Methods}      & Backbone     & Extra data & \multicolumn{1}{c|}{Epoch}   & AP$^p_{50}$ & AP$^p_{vol}$ & PCP$_{50}$ \\ \hline
Top-Down     &              &            &                               &      &       &      \\ \hline
\multicolumn{1}{l|}{M-RCNN~\cite{ZhengboZhang2022DistillingID} (TPAMI'17)}       & ResNet-50     & -          & \multicolumn{1}{c|}{-}        & 14.9 & 33.9  & 25.1 \\
\multicolumn{1}{l|}{P-RCNN~\cite{LuYang2018ParsingRF} (CVPR'18)}      & ResNeXt-101      & Pose       & \multicolumn{1}{c|}{75}      & 30.2 & 41.8  & 44.2 \\
\multicolumn{1}{l|}{M-CE2P~\cite{TaoRuan2019DevilIT} (AAAI'19)}       & ResNet-101      & -       & \multicolumn{1}{c|}{150}      & 34.5 & 42.7  & 43.7 \\
\multicolumn{1}{l|}{SNT~\cite{RuyiJi2019LearningSN} (ECCV'19)}       & ResNet-101         & -          & \multicolumn{1}{c|}{-}        & 34.4 & 42.5  & 43.5 \\
\multicolumn{1}{l|}{RP-RCNN~\cite{LuYang2020RenovatingPR} (ECCV'20)}  & ResNet-50    & Pose       & \multicolumn{1}{c|}{150}      & 45.3 & 46.8  & 43.8 \\ 
\multicolumn{1}{l|}{AIParsing~\cite{SanyiZhang2022AIParsingAI} (TIP'22)}  & ResNet-101      & -       & \multicolumn{1}{c|}{75}      & 43.2 & 46.6  & 47.3 \\ \hline
Bottom-up    &            &                               &      &       &      \\ \hline
\multicolumn{1}{l|}{PGN~\cite{KeGong2018InstancelevelHP} (ECCV'18)}   & ResNet-101  & -       & \multicolumn{1}{c|}{-}        & 17.6 & 35.5  & 26.9 \\
\multicolumn{1}{l|}{MHParser~\cite{JianshuLi2021MultihumanPW} (TOMM'21)} & ResNet-101    & -          & \multicolumn{1}{c|}{-}        & 18.0 & 36.1  & 27.0 \\
\multicolumn{1}{l|}{NAN~\cite{JianZhao2020FineGrainedMP} (IJCV'20)}  & -        & -          & \multicolumn{1}{c|}{$\sim$80} & 25.1 & 41.8  & 32.3 \\
\multicolumn{1}{l|}{DSPF~\cite{TianfeiZhou2021DifferentiableMH} (CVPR'21)} & ResNet-101  & Pose       & \multicolumn{1}{c|}{150}      & 39.0 & 44.3  & 42.3 \\ \hline
Single-stage &            &                               &      &       &      \\ \hline
\rowcolor[HTML]{E6E6E6} \multicolumn{1}{l|}{SMP~\cite{2023SMP} (arXiv preprint'23)}   & ResNet-101   & -          & \multicolumn{1}{c|}{36}       & 47.1 & 48.2  & 51.5 \\ \hline
\multicolumn{1}{l|}{UniParser$_{light}$ \rm (Ours)}  & ResNet-101   & -          & \multicolumn{1}{c|}{36}       & 51.1 & 
49.2  & 52.7 \\ 
\rowcolor[HTML]{E6E6E6} \multicolumn{1}{l|}{UniParser (Ours)}  & ResNet-101   & -          & \multicolumn{1}{c|}{36}       & \textbf{51.2} & \textbf{49.3}  & \textbf{52.9} \\ %\hline
\end{tabular}
\label{The main results on MHP}
\vspace{-3mm}
\end{table*}

\textbf{Model Runtime and Parameter Quantity Analysis.} As shown in Tab.~\ref{The runtime analysis on MHPv2.0 val with average of 2.6 people per image.}, we compare the model inference time and parameter quantity with representative open-source methods. Our method greatly reduces the inference time, which is 16\% faster than the currently fastest model, \emph{i.e.}, SMP~\cite{2023SMP}. This is due to the fact that SMP requires a post-processing step involving NMS~\cite{XinlongWang2020SOLOv2DA}, whereas our method is NMS-free.  \begin{wraptable}{r}{8.4cm}\small
    \vspace{-2.5mm}
    \caption{The model runtime and parameter quantity analysis on MHPv2.0 val set with average of 2.6 people per image. The \textcolor{blue}{blue} values indicate the decreases in inference time compared with the fastest open-source reproducible model.}
    \setlength{\tabcolsep}{3.2pt}
    \small
    \begin{tabular}{l|lccc}
    %\hline
    \multicolumn{1}{l|}{}             & Methods  & T$_{\rm infer}$~(ms) & Params (MB)  \\ \hline
    \multirow{4}{*}{Top-down}         & SNT~\cite{RuyiJi2019LearningSN}       & 3546     & -      \\
                                  & M-CE2P~\cite{TaoRuan2019DevilIT}   & 1023      & -      \\
                                  & P-RCNN~\cite{LuYang2018ParsingRF}    & 256      & 54.3      \\
                                  & RP-RCNN~\cite{LuYang2020RenovatingPR}   & 341      & 58.4      \\ \hline
    \multirow{3}{*}{Bottom-up}        & MHParser~\cite{JianshuLi2021MultihumanPW}  & 1224     & -       \\
                                  & NAN~\cite{JianZhao2020FineGrainedMP}       & 997      & -      \\
                                  & PGN~\cite{KeGong2018InstancelevelHP}     & 524        & 629.2    \\
                                  \hline
    \multirow{3}{*}{Single-stage}  & SMP~\cite{2023SMP}     & 124   & 76.21\\

    &UniParser$_{light}$ (Ours)     & 82 \textcolor{blue}{($\downarrow$ 34\%)}  & 30.54
        \\ 
    &UniParser (Ours)     & 102 \textcolor{blue}{($\downarrow$ 16\%)}  & 36.73 \\ 
   % \hline
    \end{tabular}
    \label{The runtime analysis on MHPv2.0 val with average of 2.6 people per image.}
    \vspace{-4mm}
\end{wraptable} From the perspective of model complexity, our method is not only more concise in structure but also the lightest in terms of parameter quantity.  Specifically, UniParser has 39\% fewer parameters than SMP. We can also notice that the light UniParser can further reduce about 20ms for inference time and 6.19MB model parameters with minimal performance degradation. Furthermore, compared with existing top-down methods, \emph{e.g.}, SNT~\cite{RuyiJi2019LearningSN}, P-RCNN~\cite{LuYang2018ParsingRF}, UniParser is obviously faster and more lightweight, which is even 3 times faster than the fastest method ~\cite{LuYang2018ParsingRF}. Bottom-up methods need an extra grouping process. Then they have no advantage in the inference time compared with single-stage methods. 
In particular, UniParser is about 6 times faster than PGN~\cite{KeGong2018InstancelevelHP}.

\textbf{CIHP val set.} To verify the generalization of UniParser, Tab.~\ref{The main results on CIHP} reports the results of comparison with the state-of-the-art methods on CIHP val set. We keep the same experimental settings as in MHPv2.0, with the exception of utilizing a 75-epoch training schedule as AIParsing. Finally, UniParser with ResNet-101 as backbone achieves 75.9\%, 60.4\%, and 69.0\% on AP$^{p}_{50}$, AP$^{p}_{vol}$, and PCP$_{50}$, outperforming by +0.7, +0.1, and +0.5 than the representative work AIParsing~\cite{SanyiZhang2022AIParsingAI}. %The reason why UniParser does not perform as well as in MHPv2.0 will be shown in Appendix.

\subsection{Ablation Study}

We conduct ablation experiments based on the MHPv2.0~\cite{JianZhao2018UnderstandingHI} dataset to explore the effects of various factors on the performance of UniParser. All experiments adopt a 12-epoch training schedule with UniParser whose backbone is ResNet-50~\cite{KaimingHe2015DeepRL}.

% \begin{figure}[t]
%     \centering
%     \includegraphics[width=1.0\columnwidth]{Fig/Visual_main_results.pdf}
%     \caption{Main results of UniParser. We visualize and compare the output results of our method with the best currently available model~\cite{2023SMP}.
%     }
%     %An input image is first sent into backbone with FPN to obtain multi-scale features, which serve as input for Center Locator (CL), Instance Feature Space Builder (IFSB), Category Feature Space Builder (CFSB).
%     \label{fig.Main_results}
% \end{figure}

\begin{table*}[ht]
\renewcommand\arraystretch{1.0}
\caption{The comparable results on CIHP dataset. To facilitate better comparison, we denote the state-of-the-art methods under identical settings in gray.}
\centering
\setlength{\tabcolsep}{9pt}
\small
\begin{tabular}{llccccccc}

\multicolumn{1}{l|}{Methods}      & Backbone     & Extra data & \multicolumn{1}{c|}{Epoch}   & AP$^p_{50}$ & AP$^p_{vol}$ & PCP$_{50}$ \\ \hline
Top-Down     &              &            &                               &      &       &      \\ \hline
\multicolumn{1}{l|}{BM-RCNN~\cite{2020cheng} (ECCV'20)}       & ResNet-50     & -          & \multicolumn{1}{c|}{75}        & 64.6 & 54.3  & 61.8 \\
\multicolumn{1}{l|}{P-RCNN~\cite{LuYang2018ParsingRF} 
 (CVPR'18)}      & ResNeXt-101      & Pose       & \multicolumn{1}{c|}{75}      & 69.1 & 55.9  & 66.2 \\
\multicolumn{1}{l|}{M-CE2P~\cite{TaoRuan2019DevilIT} (AAAI'19)}       & ResNet-101      & -       & \multicolumn{1}{c|}{150}      & 54.7 & 48.9  & - \\
\multicolumn{1}{l|}{SCHP~\cite{2021Self} (TPAMI'21)}      & ResNet-101   & -          & \multicolumn{1}{c|}{150} 
    & 58.9  & 52.0  & - \\
\multicolumn{1}{l|}{RP-RCNN~\cite{LuYang2020RenovatingPR} (ECCV'20)}  & ResNet-50    & Pose       & \multicolumn{1}{c|}{75}      & 71.6 & 58.3  & 62.2 \\ 
\rowcolor[HTML]{E6E6E6} \multicolumn{1}{l|}{AIParsing~\cite{SanyiZhang2022AIParsingAI} (TIP'22)}  & ResNet-101      & -       & \multicolumn{1}{c|}{75}      & 75.2 & 60.3  & 68.5 \\ 
\hline
Bottom-up    &            &                               &      &       &      \\ \hline
\multicolumn{1}{l|}{PGN~\cite{KeGong2018InstancelevelHP} (ECCV'18)}   & ResNet-101  & -       & \multicolumn{1}{c|}{-}        & 34.0 & 39.0  & 61.0 \\ \hline
Single-stage &            &                               &      &       &      \\ \hline
\rowcolor[HTML]{E6E6E6} \multicolumn{1}{l|}{UniParser (Ours)}  & ResNet-101   & -          & \multicolumn{1}{c|}{75}       & \textbf{75.9} & \textbf{60.4}  & \textbf{69.0} \\ 
\end{tabular}
\label{The main results on CIHP}
\vspace{-4mm}
\end{table*}

%\begin{figure}[th]
%    \centering
%    \includegraphics[width=1.0\columnwidth]{Fig/Similarity.pdf}
%    \caption{Visualization of cosine similarity under different conditions. (a) and (c) show the similarity matrices between category and instance kernels with the metric loss function, respectively. (b) and (d) show the similarity matrices without the metric loss function.
%    }
    %An input image is first sent into backbone with FPN to obtain multi-scale features, which serve as input for Center Locator (CL), Instance Feature Space Builder (IFSB), Category Feature Space Builder (CFSB).
%    \label{fig.similarity}
%\end{figure}

%xing：这四个表格也可以将宽度调整到linewidth，保持一致。另外，所有结果基本为表格呈现，可以更多样点，比如表5可以换成直方图形式。
\begin{table}[ht]\small
\caption{Ablation experiments.}\label{tab:ablation}  
\vspace{-1mm}
% \flushleft
\centering
\subtable[Ablation studies of the metric loss  $\mathcal{L}_{\rm metric}$. ML denotes the metric loss.]
{          
    \label{subtab:metric}
        
		\begin{tabular}{cc|ccc}
            ML & NMS & AP$^p_{50}$ & AP$^p_{vol}$ & PCP$_{50}$  \\\hline
			 &  &34.6 &43.6 &41.9 \\
             & \checkmark &34.6 &43.6 &41.9\\
             \checkmark & &42.8 &46.3 &46.7  \\
			\checkmark &  \checkmark &42.8 &46.3 &46.7  \\
		\end{tabular}
} 
\hspace{10mm} 
\subtable[The contributions for different items in the auxiliary loss $\mathcal{L}_{aux}$.]
{          
    \begin{tabular}{cc|ccc}
      %  \hline
       Ins & Cate & AP$^p_{50}$ & AP$^p_{vol}$ & PCP$_{50}$  \\\hline
			 &  &15.9 &34.6 &27.3 \\
             & \checkmark &35.5 &42.7 &27.0\\
             \checkmark & &29.4 &42.0 &38.6  \\
			\checkmark &  \checkmark &42.8 &46.3 &46.7  \\
      %  \hline
    \end{tabular}  
    \label{subtab:lossaux} 
}

\subtable[Effect of the cosine space. "Inner+Sigmoid (A)" and "(B)" denote applying sigmoid function \textbf{After} or \textbf{Before} convolution.]
{          
    \begin{tabular}{l|ccc}
     %   \hline
        Space  & AP$^p_{50}$ & AP$^p_{vol}$ & PCP$_{50}$  \\
        \hline
        %\multicolumn{13}{c}{single-scale}\\
        %\hline
        Cosine  &42.8 &46.3 &46.7 \\  
        Inner &37.6 & 42.9 & 45.1 \\
        Inner + Sigmoid (A) &0.2 & 0.1 & 0.3 \\
        Inner + Sigmoid (B) &0 & 0 & 0 \\
       % \hline
    \end{tabular}  
    \label{subtab:Space}
    \vspace{-1mm}
}   
\hspace{5mm} 
\subtable[Comparison results between different fusion methods. For detailed fusion procedure, please refer to Sec.~\ref{analysis of fusion}.]
{          
    \begin{tabular}{l|ccc}
     %   \hline
        Methods & AP$^p_{50}$ & AP$^p_{vol}$ & PCP$_{50}$  \\
        \hline
        %\multicolumn{13}{c}{single-scale}\\
        %\hline
        Index &42.8 &46.3 &46.7 \\
        Convs &32.4 & 38.4 & 34.5 \\
        Multi &37.7 & 39.2 & 38.1 \\
      %  \hline
    \end{tabular}  
    \label{subtab:fusion} 
} 
\vspace{-5mm}
\end{table}

\begin{figure}[t]
    \vspace{-4mm}
    \centering
    \includegraphics[width=1.0\columnwidth]{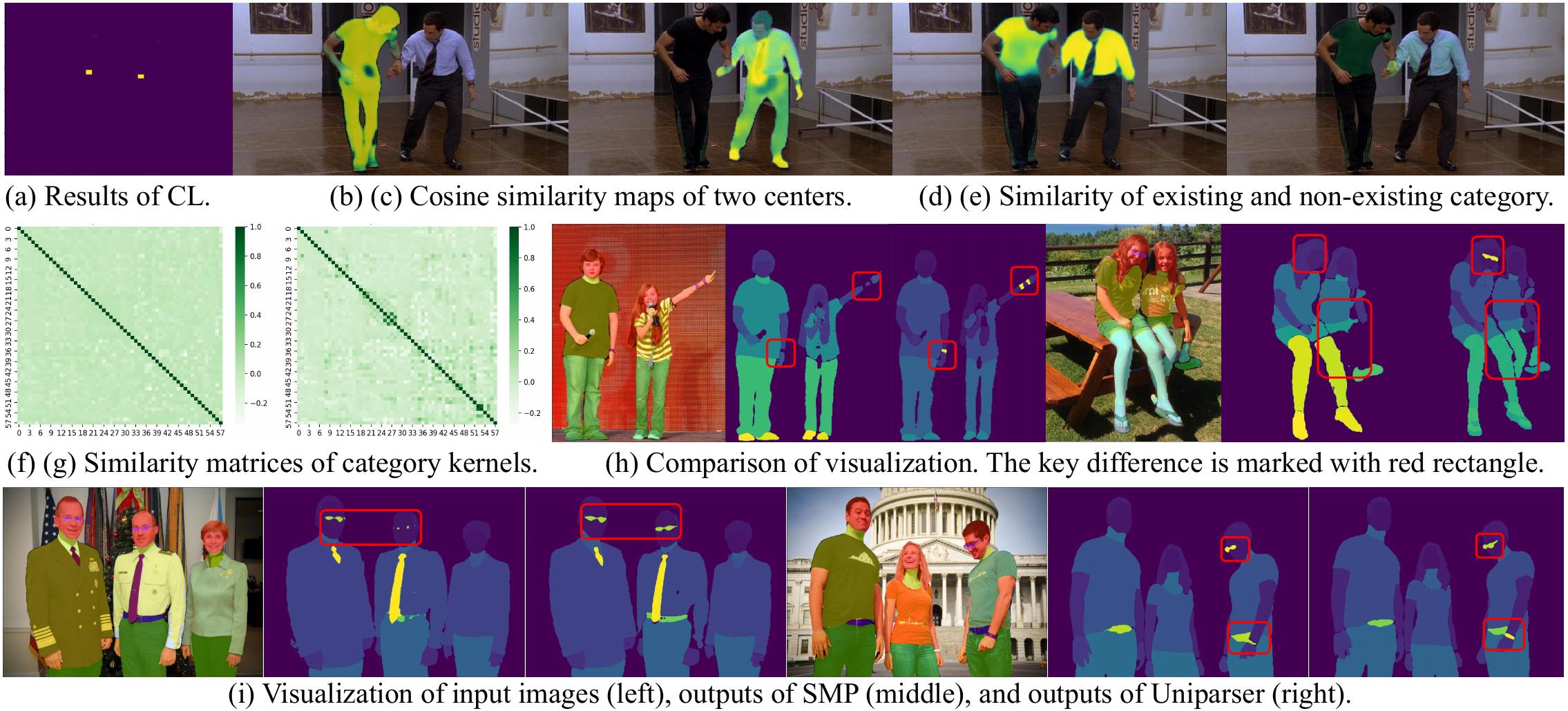}
    \vspace{-3mm}
    \caption{\textbf{The intermediate results and  visualization comparisons in UniParser.} (a) denotes the output results of CL; (b) and (c) represent the cosine similarity maps corresponding to the two centers; (d) and (e) are the similarity maps of the existing category "T-shirt" and non-existing "Polo shirt"; (f) and (g) show the similarity matrices of category kernels with and without metric loss; (h) and (i) illustrate visual comparisons between the state-of-the-art method SMP~\cite{2023SMP} and UniParser: input image (left), SMP (middle), UniParser (right). Best viewed with zoom-in.
    }    
    \label{fig.intermediate_results}
    \vspace{-2mm}
\end{figure}

\textbf{Analysis of the Metric Loss.}
The metric loss is proposed to improve the discriminative ability between instances or categories. We conduct an ablation study on the metric loss. As shown in Tab.~\ref{tab:ablation} (a), after adding the metric loss during training, the performance improvement of the model exceeds 8.2\%, 2.7\%, 4.8\% in AP$^p_{50}$, AP$^p_{vol}$, PCP$_{50}$, respectively. From Fig.~\ref{fig.intermediate_results} (f)(g), it is evident that model with the metric loss exhibits a lower cosine similarity between category convolution kernels, and displays greater dispersion in the cosine space. Thus, the metric loss effectively enhances discriminability among different instances or categories. 

\textbf{Verification of NMS-free.}
In addition, we also conduct Matrix-NMS comparative experiments to the final prediction results of UniParser. Finally, as shown in Tab.~\ref{tab:ablation} (a), whether we add the metric loss or not, NMS does not change the performance of UniParser, which illustrates that UniParser does not need the NMS process anymore. This verifies that our method is a new NMS-free method compared with the Hungarian assignment strategy~\cite{2020DETR}.

%\textbf{Discussion on NMS Process.} The objective of the NMS process is to remove redundant predictions for identical targets. NMS can be excluded when the model does not produce redundant predictions. Currently, dense prediction approaches that can achieve NMS-free typically employ Hungarian assignment algorithm~\cite{2020DETR} to establish one-to-one mapping. However, this method requires a fixed number of sufficient queries to ensure no missed detections and cannot be dynamically adjusted based on input. Different from prior methods, our method constructs a cosine space and employs self-convolution to calculate cosine similarity, which assigns each pixel exclusively to one instance and category. This effectively eliminates any overlap or redundant predictions from the headstream. 

% \begin{figure}[ht]
%     \vspace{-2mm}
%     \centering
%     \includegraphics[width=0.8\columnwidth]{Fig/Visual_main_results.pdf}
%     \vspace{-3mm}
%     \caption{The comparison between visualizations of UniParser and SOTA method SMP. Keeping good performance on large object, UniParser also can accurately segment small objects.
%     }    
%     \label{fig.visualization}
% \end{figure}

\textbf{Analysis of the Auxiliary Loss.}
The auxiliary loss function provides IFSB and CFSB with explicit training direction, which is tantamount to furnishing prior knowledge for the model: human parsing can be readily acquired based on instance features and category features. This setting enables the model to learn required features faster and better, ensuring the correct model training direction. As shown in Tab.~\ref{tab:ablation} (b), although without auxiliary loss function, $\mathcal{L}_{\rm par}$ also could drive IFSB and CFSB jointly to learn their space adaptively. Nevertheless, excluding the auxiliary loss function results in the suboptimal performance of the final model compared to its explicit supervision counterpart.

\textbf{Analysis of the Cosine Space.}
We also test other metric spaces in UniParser. Considering the design principle of using metrics to predict masks, we have chosen inner product space for comparative experiments. In contrast to cosine similarity, the magnitude of inner product similarity is affected by the length of the vectors, and there are no fixed upper or lower limits. Therefore, in addition to directly using inner product similarity as the prediction basis, we also conducted a comparative experiment that applied the sigmoid function to scale similarity. However, when applying inner product space directly, the generated mask is unstable due to the impact of the vector size. The model's training difficulty will increase significantly if the sigmoid function applies before or after self-convolution. Overall, the outcome is still inferior to cosine vector space, as demonstrated in Tab.~\ref{tab:ablation} (c).

\textbf{Analysis of Different Fusion Methods.}
\label{analysis of fusion}
%The architecture of a fusion module plays a crucial role in UniParser, as it directly impacts the quality of the final parsing results. 
From Sec.~\ref{task description}, it can be found that, given the instance label and category label, indexing the category label by the instance label can directly obtain the parsing result. However, the above operation cannot realize instance and category information exchange and fusion. Therefore, we design distinct fusion modules with comparative experiments. 
In Tab.~\ref{tab:ablation} (d), \emph{Index} is directly indexing category similarity maps with instance maps. \emph{Conv} denotes integrating category and instance features with convolution. \emph{Mulit} represents the multiplication of the similarity map with features before applying convolution layers. For detailed architecture of fusion modules, please refer to the appendix.
Seeing from the results, we found that indexing directly performs best. And we guess that additional convolution layers lead to a slower convergence process, because we need to learn the prior knowledge that instance and category information can be directly combined.
%Finally, additional convolution layers leads to a slower convergence process. Because without prior knowledge that instance and category information can be directly combined, the module must learn this relationship additionally, leading to decreased performance under identical conditions.

\textbf{Analysis of Joint Optimization.} To verify the effectiveness of joint optimization, we compare the ultimate parsing results with those achieved by directly integrating instance segmentation in IFSB and category segmentation in CFSB. And the parsing loss $\mathcal{L}_{\rm par}$ is removed. As shown in Tab.~\ref{table:joint}, it is evident that the model without $\mathcal{L}_{\rm par}$ performs inferior to the model with joint optimization, \begin{wraptable}{r}{6.5cm}\small
    \vspace{-2.5mm}
    \renewcommand\arraystretch{1.1}
    \caption{Ablation studies for joint optimization.}\label{table:joint}
	\centering
        \setlength{\tabcolsep}{7.5pt}
	\begin{tabular}{c|ccc}
	    \hline
         Loss function &  AP$^p_{50}$ & AP$^p_{vol}$ & PCP$_{50}$  \\
        \hline
        w $\mathcal{L}_{\rm par}$ &  42.8 & 46.3 & 46.7\\
        w/o $\mathcal{L}_{\rm par}$  & 30.1  & 41.5  & 34.3\\
        \hline
	\end{tabular}
 \vspace{-2mm}
\end{wraptable} which degrades 12.7\%, 4.8\%,, and 12.4\%, in AP$^p_{50}$, AP$^p_{vol}$, and PCP$_{50}$, respectively. This phenomenon demonstrates that the integration of instance and category branches through joint optimization can streamline the pipeline and improve model performance. Additionally, it facilitates the optimization of both branches with human parsing. %From Fig.~\ref{fig.intermediate_results} (g)(h), we can observe that the masks of instances and categories demonstrate a high level of quality. For more visualization results, please refer to the supplementary material.

\section{Conclusion and Future Work}
In this paper, we present UniParser, a compact and end-to-end framework for multi-human parsing. Firstly, we propose to integrate instance-level and category-level representations by fusing their features with a fusion procedure. Secondly, we introduce correlation representation learning to learn instance and category features within cosine space. Additionally, the intermediate and final network outputs for both instances and categories are standardized as pixel-level segmentation, which can be supervised by an auxiliary loss and a parsing loss respectively. Comprehensive experiments have confirmed the effectiveness and efficiency of UniParser. In the future, we aim to investigate the application of correlation representation learning in scenarios with limited image samples.

%\noindent\textbf{Limitation.} As for the limitation of UniParser, in the fusion module, we need to integrate instance and category information for each instance, which requires more memory and time resources.

\section{Appendix}

In this supplementary material, we present more details about UniParser, including: 1) The details about model architecture, training, inference, and fusion module; 2) societal impacts and limitations of our method; 3) more sensitive experiment results of parameters in UniParser; 4) more visualization results.

%训练参数
\subsection{Model Details}
In this section, we mainly provide relevant model architecture details for future reproduction. Our framework is built based on the mmdetection toolbox, and the details about model and experiment are controlled by a configuration file. We will provide a sample file at the end of this supplementary.

\textbf{Feature Extractor.} We mainly utilized ResNet-50 and ResNet-101 networks in experiments, used FPN for multi-scale feature fusion and added deformable convolution modules to enhance the ability of feature extractor. Their parameters are consistent with the conventional segmentation model settings, which can be referred in the configuration file. 

\textbf{Center Locator.} We apply the barycenter of human masks in groundtruth as the supervised labels. However, as a result of the sparsity of barycenters, there exists an imbalance between positive and negative samples within the image. Then we set a positive region based on the barycenter, which is a rectangle region centered on barycenter with shape of $(\sigma H, \sigma W)$, where $\sigma = 0.2$, $H, W$ are the height and weight of human bounding box. Each pixel falling into the region will be regarded as a positive sample.

\textbf{Instance Feature Space Builder.} The IFSB is composed of 5 convolution modules. Each convolution module is a convolution layer with a group normalization layer. The channel numbers of convolution layers are all 128. 

\textbf{Category Feature Space Builder.} The structure of CFSB is identical to that of IFSB. The category kernels are initialized through a normalization distribution with a standard deviation of 0.01.

\subsection{Training and inference Details}

\textbf{Loss function.} The weights of auxiliary loss function  $\lambda_{\rm aux}$, parsing loss function $\lambda_{\rm par}$, and metrics loss function $\lambda_{\rm metric}$ are 3.0, 3.0 and 1.0, respectively.

\textbf{Optimizer.} We choose SGD optimizer with momentum of 0.9. The learning rate varies with batch size, and each sample corresponds to a learning rate of $6.25 \times 10^{-4}$. We also apply warm-up method in first 500 iterations.

\textbf{Hyperparamenters in inference.} There are two hyperparameters during inference, the threshold to generate barycenter $\theta_{ctr} = 0.1$ and the threshold to generate masks $\theta_{masks} = 0.5$.

%融合方式对比 可以画个图分别画出三种方式不同，只画融合模块的输入+处理+输出 前面不用画
\subsection{Fusion Modules}
\begin{figure}[ht]
    \centering \includegraphics[width=1\columnwidth]{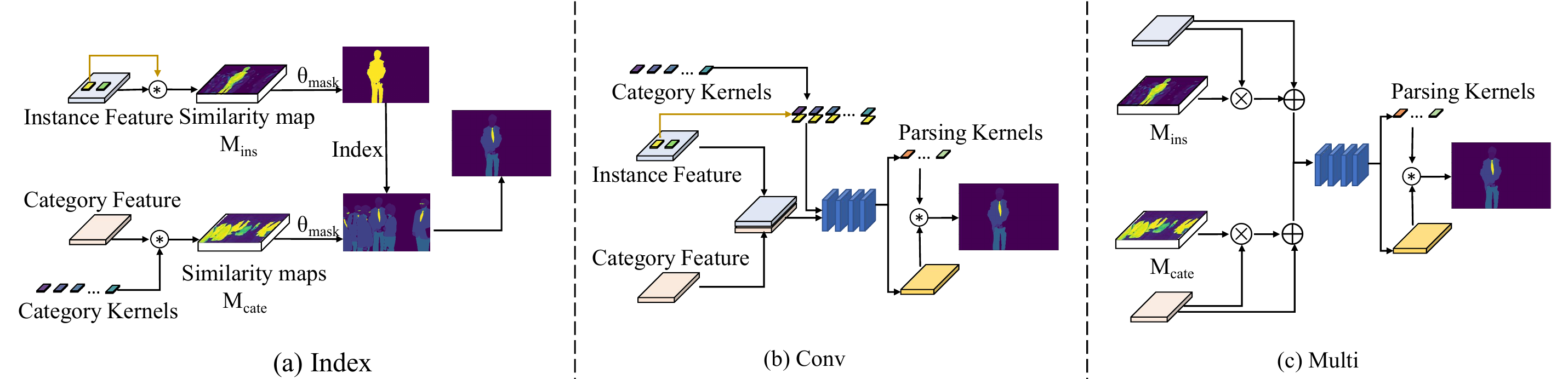}
    \caption{\textbf{Structures of different fusion modules.}}
    \vspace{-4mm}
    \label{fig.Fusion_modules}
\end{figure}

In order to achieve optimal feature integration and efficient utilization of features generated by both IFSB and CFSB, we devised three fusion modules and conducted comparative experiments. The structures of these modules are illustrated in Fig.~\ref{fig.Fusion_modules}.

%更多结果展示
\subsection{More Comparable Results.}
In this section, we present more results with different backbones and learning schedules.

\begin{table*}[ht]
\renewcommand\arraystretch{1.0}
\caption{The comparable results on MHPv2.0 dataset. To facilitate better comparison, we denote the state-of-the-art methods under identical settings in gray. $light$ refers to the reduction of convolution channels and kernel sizes in the feature compression layer after FPN.}
\vspace{-1mm}
\setlength{\tabcolsep}{8.4pt}
\small
\centering
\begin{tabular}{llccccccc}
%\hline
\multicolumn{1}{l|}{Methods}      & Backbone     & Extra data & \multicolumn{1}{c|}{Epoch}   & AP$^p_{50}$ & AP$^p_{vol}$ & PCP$_{50}$ \\ \hline
Top-Down     &              &            &                               &      &       &      \\ \hline
\multicolumn{1}{l|}{AIParsing~\cite{SanyiZhang2022AIParsingAI} (TIP'22)}  & ResNet-101      & -       & \multicolumn{1}{c|}{75}      & 43.2 & 46.6  & 47.3 \\ \hline
Bottom-up    &            &                               &      &       &      \\ \hline
\multicolumn{1}{l|}{DSPF~\cite{TianfeiZhou2021DifferentiableMH} (CVPR'21)} & ResNet-101  & Pose       & \multicolumn{1}{c|}{150}      & 39.0 & 44.3  & 42.3 \\ \hline
Single-stage &            &                               &      &       &      \\ \hline
\multicolumn{1}{l|}{SMP~\cite{2023SMP} (arXiv preprint'23)}   & ResNet-50   & -          & \multicolumn{1}{c|}{36}       & 46.9 & 48.0  & 50.1 \\ 
\rowcolor[HTML]{E6E6E6} \multicolumn{1}{l|}{SMP~\cite{2023SMP} (arXiv preprint'23)}   & ResNet-101   & -          & \multicolumn{1}{c|}{36}       & 47.1 & 48.2  & 51.5 \\ \hline
\multicolumn{1}{l|}{UniParser$_{light}$ (Ours)}  & ResNet-50   & -          & \multicolumn{1}{c|}{12}       & 41.1 & 45.6  & 45.3 \\ %\hline
\multicolumn{1}{l|}{UniParser$_{light}$ (Ours)}  & ResNet-50   & -          & \multicolumn{1}{c|}{36}       & 48.9 & 48.5  & 50.9 \\ %\hline
\multicolumn{1}{l|}{UniParser$_{light}$ \rm (Ours)}  & ResNet-101   & -          & \multicolumn{1}{c|}{36}       & 51.1 & 49.2  & 52.7 \\ \hline
\multicolumn{1}{l|}{UniParser (Ours)}  & ResNet-50   & -          & \multicolumn{1}{c|}{12}       & 42.8 & 46.3  & 46.7 \\ %\hline
\multicolumn{1}{l|}{UniParser (Ours)}  & ResNet-50   & -          & \multicolumn{1}{c|}{36}       & 49.2 & 48.7  & 51.1 \\ %\hline
\rowcolor[HTML]{E6E6E6} \multicolumn{1}{l|}{UniParser (Ours)}  & ResNet-101   & -          & \multicolumn{1}{c|}{36}       & \textbf{51.2} & \textbf{49.3}  & \textbf{52.9} \\ %\hline
\end{tabular}
\end{table*}

\begin{table*}[ht]
\renewcommand\arraystretch{1.0}
\caption{The comparable results on CIHP dataset. To facilitate better comparison, we denote the state-of-the-art methods under identical settings in gray.}
\centering
\setlength{\tabcolsep}{9pt}
\small
\begin{tabular}{llccccccc}

\multicolumn{1}{l|}{Methods}      & Backbone     & Extra data & \multicolumn{1}{c|}{Epoch}   & AP$^p_{50}$ & AP$^p_{vol}$ & PCP$_{50}$ \\ \hline
Top-Down     &              &            &                               &      &       &      \\ \hline
\rowcolor[HTML]{E6E6E6} \multicolumn{1}{l|}{AIParsing~\cite{SanyiZhang2022AIParsingAI} (TIP'22)}  & ResNet-101      & -       & \multicolumn{1}{c|}{75}      & 75.2 & 60.3  & 68.5 \\ 
\hline
Bottom-up    &            &                               &      &       &      \\ \hline
\multicolumn{1}{l|}{PGN~\cite{KeGong2018InstancelevelHP} (ECCV'18)}   & ResNet-101  & -       & \multicolumn{1}{c|}{-}        & 34.0 & 39.0  & 61.0 \\ \hline
Single-stage &            &                               &      &       &      \\ \hline
\multicolumn{1}{l|}{UniParser (Ours)}  & ResNet-50   & -          & \multicolumn{1}{c|}{75}       & 73.7 & 59.4  & 67.2 \\ 
\rowcolor[HTML]{E6E6E6} \multicolumn{1}{l|}{UniParser (Ours)}  & ResNet-101   & -          & \multicolumn{1}{c|}{75}       & \textbf{75.9} & \textbf{60.4}  & \textbf{69.0} \\ 
\end{tabular}
\label{The main results on CIHP_a}
\vspace{-4mm}
\end{table*}

%\textbf{Discussion on NMS Process.} The objective of the NMS process is to remove redundant predictions for identical targets. NMS can be excluded when the model does not produce redundant predictions. Currently, previous approaches that can achieve NMS-free typically employ Hungarian assignment algorithm~\cite{2020DETR} to establish one-to-one mapping. However, this method requires a fixed number of sufficient queries to ensure no missed detections and cannot be dynamically adjusted based on input. Different from prior methods, our method constructs a cosine space and employs self-convolution to calculate cosine similarity, which assigns each pixel exclusively to one instance and category. This effectively eliminates any overlap or redundant predictions from the headstream. 

%discussion on CIHP
\subsection{Discussion on CIHP}

From the main results in the manuscript, it can be seen that the performance improvement of the model on the CIHP dataset is much smaller than that on the MHPv2 dataset. In this section, we explain the reasons for this phenomenon.

\begin{figure}[t]
    \centering \includegraphics[width=1\columnwidth]{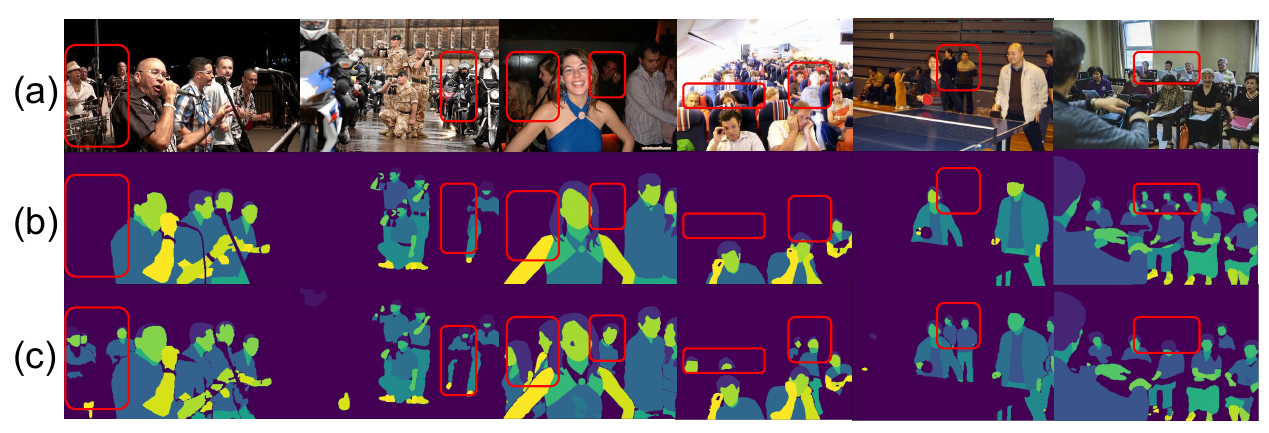}
    \caption{\textbf{The problem of missing annotations in the CIHP dataset.} (a) are the input images, (b) are the provided groundtruth, and (c) are our predictions.}
    \vspace{-4mm}
    \label{fig.missing label}
\end{figure}

\textbf{The Problem of Missing Annotations in the Dataset.}
In the CIHP dataset, there are numerous instances of missing annotations, particularly evident in images with multiple individuals where only a few are annotated. This means that the model will regard unlabeled people as the background, which has a significant negative impact on bottom-up and single-stage methods that learn category knowledge with full image. As illustrated in Fig.~\ref{fig.missing label}, missing annotations in the dataset are extremely common in multi-person images, and their occurrence is random and somewhat influenced by instance brightness and scale. However, there is no discernible pattern that can be learned. For instance, in the last two images, some small-scale occluded individuals are not labeled in the penultimate image but are labeled in the last image. This situation significantly impacts Center Locator's ability to learn small-scale people. In addition, both the bottom-up approach and our CFSB require back propagation of all pixel features belonging to the same category in the image. However, due to the issue of missing annotations, many pixels in other categories are considered as background class, which have a negative impact on the distribution learning of category semantics. For the top-down approach, their semantic branch only learns the pixels within the annotated boxes, so it is not affected by the error distribution.

\textbf{Mismatch between Test Metrics and the Dataset.}
 In addition to training, the absence of annotations can compromise the accuracy of model learning and negatively impact the testing process. Similar to the MHPv2.0 dataset, we chose AP$^{p}_{50}$, AP$^{p}_{vol}$, and PCP as the evaluation metrics. In multi-human tasks, the above metrics have a characteristic that redundant instance predictions are added to the final average calculation as erroneous predictions with an IOU value of 0. Therefore, even if individuals do exist within the red box depicted in Fig.~\ref{fig.missing label} and their category prediction is accurate, they will still be deemed incorrect predictions, thereby significantly reducing average precision. 

\textbf{Mitigation Methods.}
Due to the aforementioned characteristics, we have implemented corresponding mitigation methods that may not entirely resolve the aforementioned issues but have enhanced UniParser's performance on the CIHP dataset. Firstly, in order to reduce the impact of category branch learning on the distribution of error semantics, we eliminated the negative sample loss function in the auxiliary loss and introduced a new loss function that brings pixel features belonging to the same category closer together.

\subsection{Societal Impacts and Limitations}
\textbf{Societal Impacts.} Our work introduces a unified approach for instance-aware multi-human parsing. The work is mainly evaluated on open-sourced datasets, \emph{i.e.}, MHPv2.0 and CIHP. Besides, we follow the MHPv2.0 and CIHP terms of usage. We build on top of open source projects respecting licenses, and release all code, trained models and datasets for research purposes. Currently, the application of this approach is
mainly limited to virtual reality. Although our results on MHPv2.0 and CIHP are promising, the immediate broader impact of our work in real-world scenarios, beyond the research community, is limited. But, we believe that UniParser, as a lightweight and efficient parsing tool, reduces the application threshold for multi human parsing tasks and lays the foundation for the emergence of future fine grained vision task applications.

\textbf{Limitations.} As for the limitations of UniParser, in the fusion module, we require integration of instance and category information for each instance, which necessitates more memory and time resources. Additionally, while UniParser has been verified only in fine-grained human parsing, its performance on general semantic segmentation or instance segmentation tasks remains untested. We aim to extend our framework to other tasks and address these issues in future works. 

\subsection{Sensitive Experiments}
In this part, we conduct more experiments to analyze the effect of different parameters in UniParser. All experiments are conducted on a 12-epoch training schedule, based on ResNet-50 backbone, trained with MHPv2.0 dataset. Uniparser has fewer hyperparameter. The only adjustable hyperparameter involved in the training process are the size of the feature grids in the Center Locator and the coefficient $\sigma$ to generate labels. We conducted relevant experiments to compare the performance of the model under different feature grids sizes and different $\sigma$ in Tab.~\ref{grid experiments_a.}.

\begin{table}[ht]
\caption{The impact of grid numbers and center region on MHPv2.0 val set. 0.5$\times$, 2$\times$ denote the grid numbers multiplied by the corresponding coefficients. $\sigma$ is the hyperparameter controlling the scale of center region. FPS indicates the inference speed on RTX3090.}
\centering
\setlength{\tabcolsep}{11.6pt}
\small
\begin{tabular}{cc|cccc}

Grids & $\sigma$& AP$^{p}_{50}$   & AP$^{p}_{vol}$ & PCP$_{50}$ &FPS \\ \hline
0.5$\times$  & 0.2   & 33.6 & 39.1    & 38.7  &13.1  \\
0.5$\times$ & 0.1   & 29.5 & 36.2   & 33.2  &13.1  \\ \hline
 1$\times$    & 0.2   &  42.8 & 46.3 & 46.7  &12.2  \\
 1$\times$    & 0.1   & 41.1 & 45.7    & 45.5  &12.2  \\ \hline
 2$\times$    & 0.2   & 41.5 & 46.0    & 46.3  &11.2  \\
 2$\times$    & 0.1   & 40.8 & 45.2    & 44.9  &11.2  \\ 
\end{tabular}
\captionsetup{font={small}}
\label{grid experiments_a.}

\end{table}

Due to the $\sigma$ and number of grids only affect the positioning effect of the Center Locator in UniParser, so blindly increasing its resolution will not bring any performance improvement.

In addition, we also conducted sensitive experiments on two hyperparameter, involved in the inference process, $\theta_{masks}$ and $\theta_{ctr}$. $\theta_{ctr}$ is used to filter the prediction results of Center Locator, and the predicted position with a confidence greater than $\theta_{ctr}$ will be considered as the instance barycenter. $\theta_{masks}$ is used to ultimately generate fine masks. When the pixel similarity in output results is greater than the $\theta_{masks}$, the pixel is considered as the foreground.

\begin{table}[ht]

\caption{The impact of two thresholds on MHPv2.0 val set. }
\centering
\setlength{\tabcolsep}{11.6pt}
\small
\begin{tabular}{cc|cccc}

$\theta_{masks}$  & $\theta_{ctr}$& AP$^{p}_{50}$   & AP$^{p}_{vol}$ & PCP$_{50}$\\ \hline
0.5  & 0.2   &  42.6 & 46.2 & 46.6   \\
0.5  & 0.1   & 42.8 & 46.3 & 46.7  \\ 
0.5  & 0.05   & 42.5 & 46.2 & 46.5   \\ \hline
0.45 & 0.2   &  42.4 & 46.0 & 46.2  \\
0.45 & 0.1   & 42.6 & 46.1 & 46.4  \\
0.45 & 0.05   & 42.4 & 45.9 & 46.0  \\ \hline
0.55 & 0.2   & 42.3 & 45.9 & 46.3  \\
0.55 & 0.1   & 42.6 & 46.1 & 46.4  \\ 
0.55 & 0.05   & 42.2 & 45.9 & 46.1  \\ 
\end{tabular}
\captionsetup{font={small}}
\label{grid experiments.}

\end{table}

\subsection{More Visualization Results}

% 实例级 / 语义级 / 细粒度的
\subsubsection{Visualization on MHPv2.0}
We present additional visualization results on MHPv2.0, including parsing results, instance-level segmentation results, and category-level segmentation results. We can find that UniParser can perform well in various scenes.

\begin{figure}[ht]
    \centering \includegraphics[width=1\columnwidth]{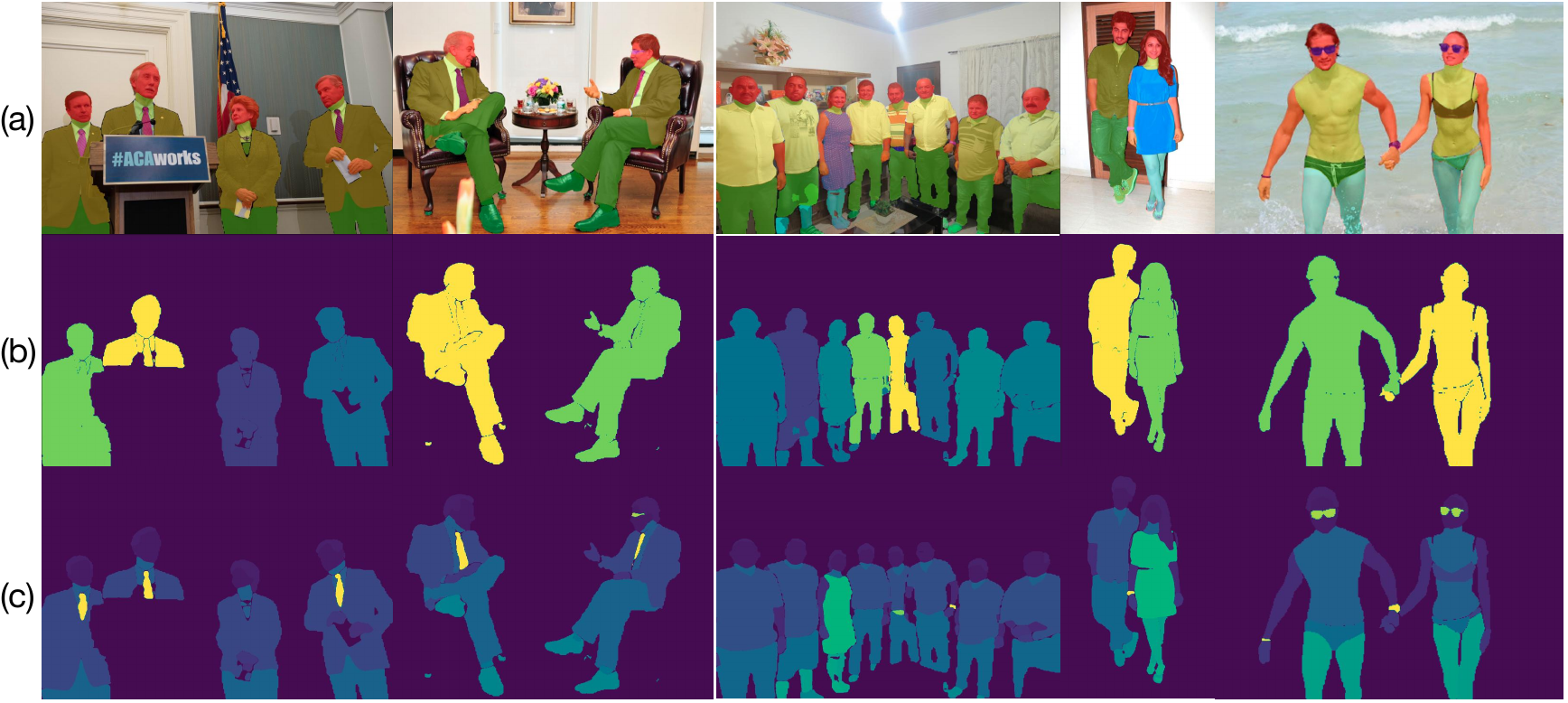}
    \caption{\textbf{Visualization results on MHPv2.0.} (a) are the human parsing results, (b) are the instance-level results, (c) are the category-level results.}
    \label{fig.uniparser_sup2}
\end{figure}

\subsubsection{Visualization on CIHP}
We further present visualization results on CIHP to verify the generalization of UniParser.

\begin{figure}[ht]
    \centering \includegraphics[width=1\columnwidth]{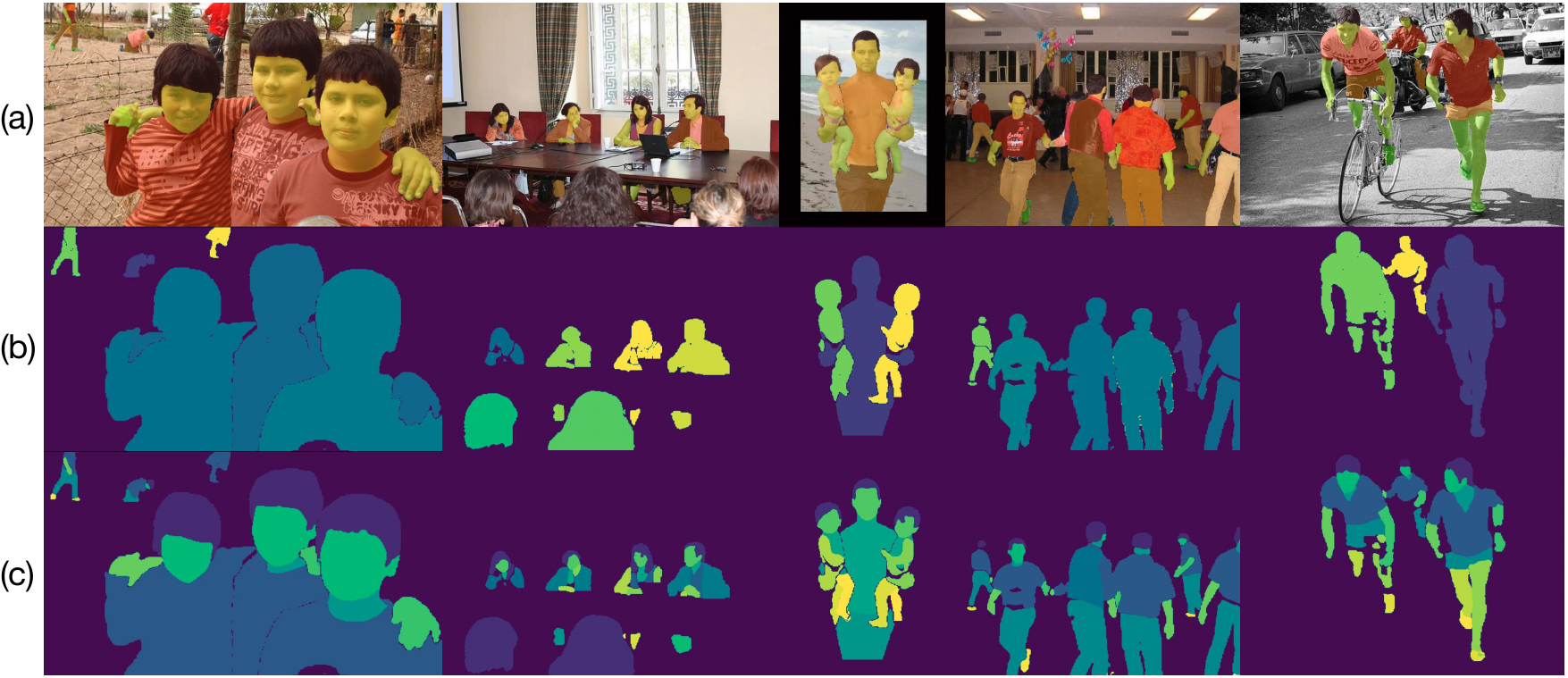}
    \caption{\textbf{Visualization results on CIHP.} (a) are the human parsing results, (b) are the instance-level results, (c) are the category-level results.}
    \vspace{-4mm}
    \label{fig.uniparser_sup2(1)}
\end{figure}

\newpage

%Xing: 参考文献格式需要统一，包括人名、篇名、题名等是否大小写及缩写一致性，不同类型文献的包含条目一致性，发表名一致性等。
\medskip
{
\small
\bibliographystyle{unsrt}
\bibliography{Uniparser}
}

\end{document}